\pdfoutput=1
\documentclass[10pt,letterpaper,twocolumn]{article}
\usepackage[margin=0.75in,columnsep=0.25in]{geometry}
\usepackage[hyphens]{url}
\usepackage{graphicx}
\usepackage{amsmath}
\usepackage{amssymb}
\usepackage{booktabs}
\usepackage{natbib}
\usepackage{caption}
\usepackage{authblk}
\usepackage{microtype}
\usepackage[hidelinks]{hyperref}
\title{When Derived Measurements Mislead: Quantifying and Mitigating LLM Over-Trust with Privileged-Modality Reliability Evidence}
\author[1]{Zongheng Guo}
\author[2]{Tao Chen}
\author[5]{Tianli Li}
\author[2]{Mingzhe Cui}
\author[3]{Yang Jiao}
\author[2]{Lei Xie}
\author[3]{Yi Pan}
\author[4]{Xiao Hu}
\author[1]{Manuela Ferrario}
\affil[1]{Department of Electronics, Information and Bioengineering, Politecnico di Milano, Milan, Italy}
\affil[2]{State Key Laboratory of Industrial Control Technology, Zhejiang University, Hangzhou, China}
\affil[3]{Shenzhen Institutes of Advanced Technology, Chinese Academy of Sciences, Shenzhen, China}
\affil[4]{Nell Hodgson Woodruff School of Nursing, Emory University, Atlanta, USA}
\affil[5]{Department of Integrated Traditional Chinese and Western Medicine Cardiology, China-Japan Friendship Hospital, Beijing, China}

\date{}

\begin{document}
\maketitle

\begin{figure*}[!t]
\centering
\includegraphics[width=0.98\textwidth]{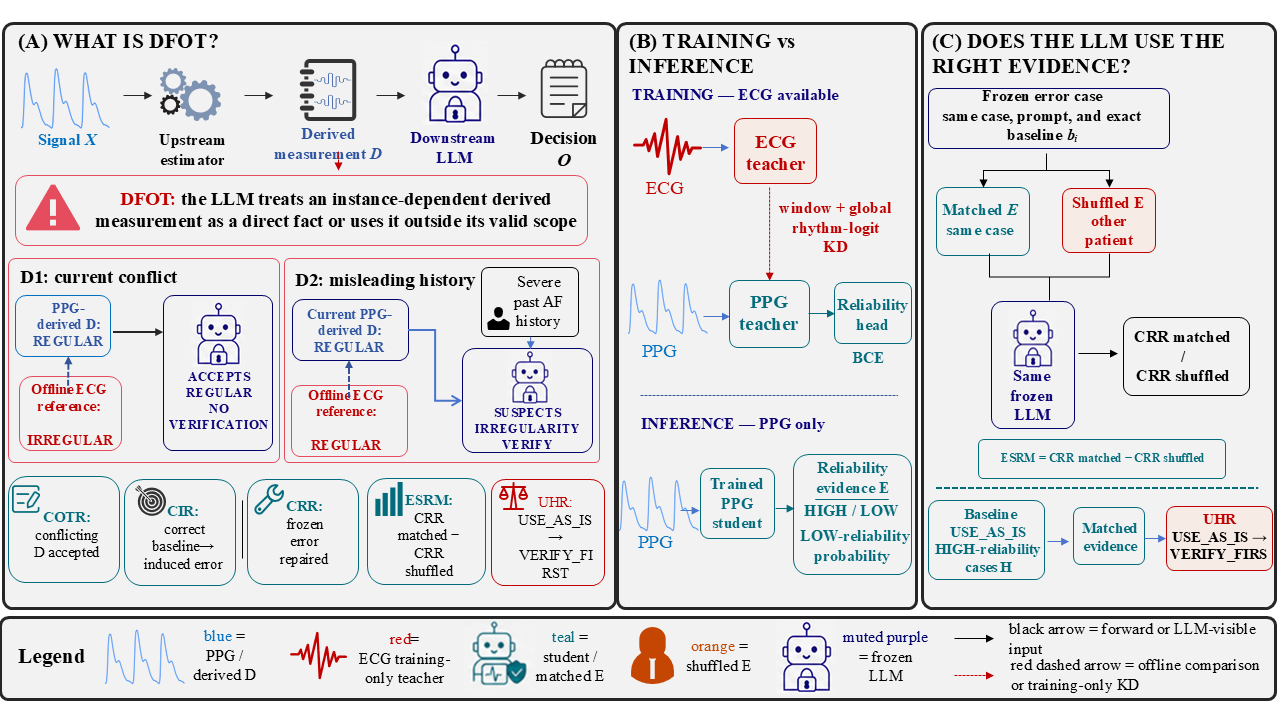}
\caption{End-to-end instantiation of the DFOT framework. A proof-of-concept privileged student converts training-only ECG supervision into PPG-only reliability evidence. The baseline reliability generator (B2) uses a conventional PPG-only reliability model, whereas the proposed privileged generator (K2) distills ECG-derived privileged information into the deployment-time PPG reliability estimate. The central matched-versus-shuffled evaluation measures repair (CRR) and evidence-package specificity (ESRM) at the LLM interface. KD: Knowledge Distillation; AF: atrial fibrillation; E: reliability evidence.}
\label{fig:overview}
\end{figure*}

\begin{abstract}
Derived measurements increasingly enter large language model (LLM) pipelines as direct facts despite their instance-dependent validity. We define \emph{derived-feature over-trust} (DFOT) as the failure in which a downstream LLM assigns such a measurement the epistemic status of a direct fact or uses it outside its valid scope. Using physiological sensing as a case study, \textbf{D1} tests acceptance of a PPG-derived rhythm contradicted by offline ECG, whereas \textbf{D2} tests rejection of an offline-confirmed reliable PPG rhythm under misleading severe history. ECG supplies training supervision and offline reference construction but is never shown to the LLM. Five estimands quantify this chain: conflict over-trust rate (COTR) and context-induced error rate (CIR) characterize D1/D2; correct repair rate (CRR) measures frozen-error repair; evidence-specific repair margin (ESRM) contrasts matched and patient-disjoint shuffled evidence; and utility harm rate (UHR) measures unnecessary verification among HIGH-reliability cases used without verification at baseline. The framework does not depend on a particular reliability generator. We demonstrate it on 50,000 paired PPG--ECG records using ECG-to-PPG privileged distillation as an illustrative baseline and PPG-only inference. On a protocol-locked 187-patient test, the baseline improves four repair and specificity endpoints by 1.82--6.69 percentage points, with all paired confidence intervals excluding zero; UHR increases by 0.67 percentage points (95\% CI: $-0.4$ to $+1.7$). DFOT provides a common evaluation target for stronger mitigation methods. The code is available at: \url{https://github.com/Zongheng-Guo/When-Derived-Measurements-Mislead}.
\end{abstract}

\section{Introduction}

Many contemporary AI systems follow a modular architecture in which specialized upstream models first convert raw observations into structured intermediate representations, and a downstream large language model (LLM) subsequently reasons over these representations to answer questions or recommend actions.
First, an upstream model maps raw observations $X=(x_1,..., x_N)$ to a compact representation or derived measurements $D=g(X)$, then a downstream large language model (LLM) uses $D$ to answer a question or recommend an action. Such architectures are increasingly common in vision systems that connect LLMs to specialized visual models or computer-aided diagnosis outputs \cite{wang2024chatcad,wu2023visualchatgpt}, in audio systems that coordinate speech interfaces and specialized audio models \cite{huang2024audiogpt}, and in signal-language systems that pass physiological summaries, derived insights, or encoded time series representations into language reasoning \cite{fang2024physiollm,langer2025opentslm}. Although this decomposition improves modularity and accessibility, it can transform an uncertain estimation process into an apparently factual token or field. In many deployments, the downstream LLM only observes $D$. It has no access to the original measurements, the acquisition conditions, or evidence about whether the upstream estimate is reliable for the current instance.

This results in a type of failure that is different from both hallucinations and conventional prediction errors. Even if a  downstream model reasons coherently from the information it receives, it can still be wrong because it assigns a derived estimate the epistemic status of a direct fact. We call this \emph{derived-feature over-trust} (DFOT). DFOT can arise whenever an LLM uses an upstream-derived value whose validity depends on the specific instance. Therefore, it is an interface-level problem rather than a pathology specific to one modality. 
Unlike probability calibration, which concerns whether an upstream model's confidence reflects empirical correctness \cite{guo2017calibration}, DFOT concerns a downstream decision maker relies on a derived value appropriately for the specific instance. Unlike selective prediction \cite{geifman2017selective,geifman2019selectivenet}, the uncertainty originates in an upstream measurement channel and must survive a semantic interface to a different model. In high-stakes settings such as medicine, failure at this interface can cause an appropriate upstream estimate to be used inappropriately downstream.

LLM studies have shown that they exhibit unstable behaviour when contextual and parametric evidence conflict \cite{longpre2021conflicts,chen2022rich,wu2024clasheval,cattan2025dragged}. Several studies suggest that medical LLMs often produce definitive answers even under substantial uncertainty, rather than abstaining or expressing calibrated confidence \cite{singhal2023medpalm,cocchieri2026abstain}. However, existing evaluations usually treat context as something to be accepted or rejected. They do not directly characterize whether an LLM appropriately relies on a value produced by another estimator when that value is valid for some instances but misleading for others. The distinction is crucial: access to a measurement doesn't mean that the downstream model uses it appropriately for each instance.

We use physiological sensing as a concrete and auditable case study. In the field of wearable health technology, photoplethysmography (PPG) data is converted into heart rate or rhythm labels,  which are then interpreted by a language model. Although PPG is convenient and widely deployed, its reliability can vary depending on rhythm, motion, perfusion, device, and acquisition conditions \cite{allen2007ppg,charlton2023roadmap}. The uncertainty introduced during this estimation process is largely hidden once only the derived feature is presented to the downstream model. In our implementation, $D$ contains a PPG-derived heart rate and categorical rhythm assessment, while a synchronized ECG serves as a higher-fidelity offline reference. This enables two controlled DFOT scenarios. In \textbf{D1}, the LLM receives a PPG-derived rhythm label that is contradicted by the synchronized ECG reference. In \textbf{D2}, misleading historical information encourages the LLM to reject a current PPG-derived rhythm that is supported by agreement between PPG and ECG. The ECG is only used for privileged training and offline evaluation, and is never shown to the downstream LLM.

Rather than proposing another reliability estimator, we ask how over-reliance on derived measurements should be defined, induced, quantified, and distinguished from generic caution. The privileged ECG signal provides one concrete case study, but the primary object of study is the downstream failure itself and the quantities used to characterize it. Our contributions are:
\begin{itemize}
    \item We introduce DFOT as a distinct downstream failure mode: an LLM assigns a derived estimate the epistemic status of a direct fact, rather than treating it as an uncertain measurement whose validity depends on the current instance.
    \item We formalize two complementary challenge mechanisms and five linked estimands---COTR, CIR, CRR, ESRM, and UHR---that use fixed (``frozen'') denominators to disentangle failure induction, conditional repair, evidence specificity, and interface harm.
    \item We introduce a matched-versus-patient-disjoint-shuffled evidence intervention that tests whether revision depends on information specific to the current instance rather than the mere presence of reliability language.
    \item We provide a protocol-locked PPG--ECG benchmark together with a privileged-distillation baseline, establishing a reproducible point of comparison for stronger future DFOT mitigation methods.
\end{itemize}

\section{Related Work}

\paragraph{Appropriate reliance and evidence use.}
Knowledge-conflict benchmarks manipulate retrieved evidence or introduce conflicts between contextual evidence and a model's parametric knowledge \cite{longpre2021conflicts,chen2022rich,wu2024clasheval}. Self-evaluation and semantic entropy estimate the reliability or uncertainty of generated answers \cite{kadavath2022know,kuhn2023semantic}. 
Beyond question answering, LLMs increasingly serve as decision-making agents in domains such as autonomous scientific analysis and code optimization \cite{yuan2026sp, kuang2025learning}. These settings further highlight the importance of appropriately using intermediate information, although robust iterative optimization remains challenging \cite{nie2026understanding}. 
Calibration, selective prediction, and learning to defer address predictive uncertainty and decision risk \cite{guo2017calibration,lakshminarayanan2017ensembles,ovadia2019uncertainty,geifman2019selectivenet,madras2018defer,mozannar2020defer}. In contrast, DFOT asks whether a downstream agent appropriately relies on an instance-specific derived measurement. However, warnings, explanations, or attention weights do not demonstrate that a model uses evidence appropriately\cite{jain2019attention,wiegreffe2019attention}. We therefore evaluate evidence use through matched--shuffled interventions and decision-level estimands.

\paragraph{Derived physiological measurements.}
PPG enables scalable heart-rate and rhythm monitoring \cite{tison2018af,perez2019apple}, but error varies across devices and activities \cite{shcherbina2017accuracy,bent2020inaccuracy}. Raw-waveform, multitask, signal-quality-aware, and foundation models improve physiological representations \cite{torressoto2020deepbeat,aschbacher2020raw,antiperovitch2024continuous,pereira2020review,pillai2025papagei,guo2026sigma,guo2025qualityfm}. Related works also investigate multimodal transfer, remote sensing, and learning from noisy physiological signals \cite{geenjaar2026robust,guo2023remotehr,chen2024actnet,chen2026chirprate,li2026biox,tang2024bp,langer2025opentslm}. These works improve an upstream estimator, whereas DFOT evaluates whether a downstream LLM appropriately relies on the resulting derived measurements.

\paragraph{Privileged information as an instantiation.}
Learning with privileged information (LUPI) and distillation transfer supervision available only during training  \cite{vapnik2009lupi,hinton2015distilling,lopezpaz2016unifying}, including ranking-based, imperfect-teacher, cross-modal, and ECG-guided PPG formulations \cite{yang2022privilegedranking,martinezgarcia2025tpd,gupta2016crossmodal,wei2025ecgppg,ni2025ppgdistill}. We use this family of methods to construct one reliability signal, but  our primary contribution is the formulation of DFOT as a downstream failure mode together with controlled interventions and estimands that quantify failure induction, repair, evidence specificity, and downstream harm.

\section{DFOT: problem formulation and metric chain}

In general, let $X_{d,i}=(x_{d,i,1},\ldots,x_{d,i,N_i})$ denote the deployment-modality measurements available for instance $i$, and let $X_{p,i}=(x_{p,i,1},\ldots,x_{p,i,T_{p,i}})$ denote a synchronized, higher-fidelity privileged modality available only during development and offline evaluation. The two modalities may be multivariate and need not have identical sampling rates or sequence lengths, provided that they refer to the same physiological episode.
An upstream estimator maps the deployment signal to a derived feature, $D_i=g(X_{d,i})$, while a separate deployment-time reliability model produces a reliability estimate, $R_i=S(X_{d,i})$. The reliability estimate may be converted into semantic evidence through a rendering function $h$ so that $E_i=h(R_i)$.
The downstream model $M$ receives a task description $Q_i$, contextual information $C_i$, the derived feature $D_i$, and, optionally, the reliability evidence $E_i$:
\begin{equation}
  O_i=M(Q_i,D_i,C_i,E_i), \qquad X_{p,i}\notin \text{input}(M).
\end{equation}
Here, $E_i=\varnothing$ denotes the no-evidence condition. The privileged modality is never provided to the downstream model.

Derived-feature over-trust occurs when the downstream model fails to condition its reliance on $D_i$ on the feature's instance-specific validity or intended scope. In D1, the model accepts a current derived feature that is contradicted by the privileged reference. In D2, out-of-scope historical context causes the model to reject a current derived feature that is supported by agreement between the deployment and privileged modalities. In our instantiation, ECG is treated as a higher-fidelity rhythm reference rather than as infallible ground truth.

Let $\mathcal{C}$ denote the set of reference-confirmed D1 conflict instances, and let $\mathcal{N}$ denote the set of D2 instances for which the no-evidence, neutral-context condition yields the correct downstream decision. Let $b_i$ denote the parsed downstream decision in the corresponding
no-evidence challenge condition, and let $y_i$ denote the parser-defined
target decision. Let $\mathcal F$ denote the relevant challenge-specific
frozen error set, $\mathcal F_1$ for D1 or $\mathcal F_2$ for D2, as defined
below. In our experiments, $y_i=
\begin{cases}
Suspect, & i\in\mathrm{D1},\\
Trustworthy, & i\in\mathrm{D2}.
\end{cases}$.

Let $d_i^{\mathrm{severe}}$ denote the parsed decision in the no-evidence
severe-history condition. Let $d_i(e)$ denote the parsed downstream decision
under evidence condition $e$, and let $u_i(e)$ denote its parsed
measurement-use field. Let
$z_i\in\{\text{\textsc{High}},\text{\textsc{Low}}\}$ denote the offline
reference-defined reliability class, which is distinct from the
deployment-time estimate $R_i$. The baseline-eligible utility set is
\[
\mathcal H=
\left\{
i:z_i=\text{\textsc{High}}
\ \land\
u_i(\varnothing)=\text{\textsc{Use-As-Is}}
\right\}.
\]
We abbreviate matched and shuffled evidence by $m$ and $s$. The five estimands are
\begin{equation}
\label{eq:dfot-metrics}
\begin{aligned}
\mathrm{COTR}
 &= \frac{1}{|\mathcal C|}\sum_{i\in\mathcal C}
    \mathbf{1}\!\left\{b_i\text{ accepts }D_i\right\}, \\[2pt]
\mathrm{CIR}
 &= \frac{1}{|\mathcal N|}\sum_{i\in\mathcal N}
    \mathbf{1}\!\left\{d_i^{\mathrm{severe}}\neq y_i\right\}, \\[2pt]
\mathrm{CRR}(e)
 &= \frac{1}{|\mathcal F|}\sum_{i\in\mathcal F}
    \mathbf{1}\!\left\{d_i(e)=y_i\right\}, \\[2pt]
\mathrm{ESRM}
 &= \mathrm{CRR}(m)-\mathrm{CRR}(s) \\
 &= \frac{1}{|\mathcal F|}\sum_{i\in\mathcal F}
    \Bigl[
      \mathbf{1}\!\left\{d_i(m)=y_i\right\}
      -\mathbf{1}\!\left\{d_i(s)=y_i\right\}
    \Bigr], \\[2pt]
\mathrm{UHR}
 &= \frac{1}{|\mathcal H|}\sum_{i\in\mathcal H}
    \mathbf{1}\!\left\{
      u_i(m)=\text{\textsc{Verify-First}}
    \right\}.
\end{aligned}
\end{equation}
COTR measures acceptance of a derived measurement despite conflict with the privileged reference , CIR measures context-induced rejection of a derived measurement supported by the privileged reference, and CRR measures parser-defined decision repair on a frozen error set. ESRM contrasts matched and shuffled evidence packages beyond mere presence of reliability evidence. UHR intentionally focuses on evidence-induced unnecessary verification of initially usable measurements. Together, these estimands transform DFOT from an informal concern into a set of falsifiable comparisons with explicitly defined denominators.

\paragraph{Frozen estimands.}
D1 freezes
$\mathcal F_1=\{i\in\mathcal C:b_i\neq y_i\}$,
whereas D2 first restricts to $\mathcal N$ and freezes
$\mathcal F_2=\{i\in\mathcal N:d_i^{\mathrm{severe}}\neq y_i\}$.
All evidence conditions use these fixed denominators. ESRM operationally
measures matched-package specificity beyond the generic effect of receiving a
reliability channel; it does not separately identify label, score, or wording
effects.

\paragraph{Claim hierarchy.}
The design separates four progressively stronger claims:
\emph{accessibility}, whether a reference channel contains learnable
information; \emph{transfer}, whether it improves a deployment-time reliability
signal; \emph{utilization}, whether that signal repairs downstream decisions;
and \emph{specificity}, whether repair depends on the current instance.
Upstream accuracy addresses the first two claims, CRR addresses utilization,
and ESRM is required for specificity.

\paragraph{Benchmark and reporting contract.}
The protocol is agnostic to the reliability generator: calibration, ensembles, selective prediction, multimodal teachers, or other uncertainty models may provide case-level evidence. A valid DFOT instantiation requires an independently defined reference relation that determines whether reliance on $D_i$ is appropriate, an observable reliance failure, and a downstream target that can be evaluated independently of the reliability signal. The evidence must not reveal the withheld reference value. A matched--control intervention must preserve the case, task, evidence format, and decision history while breaking only the correspondence between the evidence and the current instance. Cases, baseline decisions, parsers, and denominators must be frozen before comparison. Each challenge should report its baseline characterization rate, absolute matched and shuffled outcomes, their paired difference, and uncertainty respecting the sampling unit. Absolute CRR reveals residual error burden, ESRM distinguishes matched-case evidence use from generalized caution, and UHR or a domain-appropriate cost endpoint quantifies harm on a separately defined set whose baseline decision is already useful. We avoid a weighted composite because acceptable tradeoffs among repair, specificity, abstention, workload, delay, and missed detection are application dependent; the metric vector instead supports auditable Pareto comparisons.

\section{DFOT evaluation protocol}

Using the method-independent contract above, we instantiate two complementary challenge mechanisms. D1 presents hidden-irregularity cases in which the PPG-derived rhythm appears regular and freezes the subset that the downstream LLM accepts despite disagreement with the offline ECG reference. D2 holds a currently regular segment fixed while introducing severe atrial fibrillation (AF)-related history that induces an incorrect judgment of the current window. Only cases that are answered correctly under neutral history are eligible for the frozen baseline-error set.

\paragraph{Challenge sets and population estimates.}
D1 and D2 are controlled stress tests rather than prevalence estimates. Each holds the target case fixed, isolates one reliance mechanism. COTR and CIR verify that the intended failure has been induced before CRR and ESRM evaluate conditional repair on the corresponding frozen denominator. These rates therefore characterize DFOT inducibility and conditional repair, not population event frequency.

\paragraph{Matched, shuffled, and harm controls.}
Reactive mitigation appends reliability evidence to the frozen baseline interaction, directly testing revision of an observed error. Matched evidence is taken from the current record. Patient-disjoint shuffled evidence preserves the case, baseline answer, timing, treatment arm, Student seed, and prompt template while replacing only the evidence source.
The resulting matched--shuffled CRR difference defines evidence-package specificity through ESRM. A separate baseline-eligible HIGH-reliability set $\mathcal{H}$ measures UHR under matched evidence: unnecessary movement from \textsc{Use-As-Is} to \textsc{Verify-First}.

\paragraph{Frozen execution and inference.}
E0 evaluates Qwen3-8B under the primary P1 prompt; E1 holds P1 fixed across Qwen3-8B, DeepSeek-V4-Pro, and GPT-5.5; E2 tests three semantically equivalent prompts within Qwen3-8B; and E3 applies the frozen Qwen3-8B/P1 protocol to patient-disjoint test patients with no prior downstream LLM queries. Qwen requests use vLLM 0.25.1, temperature 0, at most 96 output tokens, disabled thinking, and deterministic parsing. Before E3 inference, cases, prompts, parser, donor rules, thresholds, gates, and bootstrap settings were frozen. E3 uses 5,000 patient-cluster bootstrap replicates, resampling three Student seeds with replacement and sharing each seed draw across paired arms and evidence conditions. Four efficacy gates require the lower confidence bounds for the differences in matched CRR and ESRM between the proposed K2 generator and the baseline B2 generator to remain positive in both D1 and D2 (see Figure~\ref{fig:overview} (B)).

\section{Proof-of-Concept Case Study: ECG-Referenced PPG}

The empirical case study is designed to validate the DFOT formulation's ability to distinguish reliability interfaces, rather than propose an exhaustive solution. ECG-referenced supervision provides a specific proof-of-concept signal that is not available at deployment. The evaluation framework itself does not depend on this particular teacher, modality pair, or distillation objective.

\paragraph{Cohort and cases.}
The source scan of the MIMIC-III Matched Waveform Database v1.0 identified 1,297 candidate patients with simultaneous PLETH and Lead-II recordings. After segmentation and an eligibility check, the final cohort comprised 50,000 synchronized four-minute records from 1,275 patients: 36,115 records from 925 patients (pts) were used for training, 6,464 records form 163 pts for validation, and 7,421 records from 187 pts for a locked test. These splits are patient-disjoint. PPG was resampled at 50 Hz (12,000 samples per each segment) and Lead-II ECG at 125 Hz (30,000 samples per segment). Reliability is defined based on four clear rhythm cases: both regular and both irregular are HIGH, while PPG-regular/ECG-irregular (hidden irregularity) and PPG-irregular/ECG-regular (false alarm) are LOW. The test contains 3,733 clear cases (897 LOW, 2,836 HIGH), including 401 cases of hidden irregularity and 2,151 cases of both regular rhythms.

This construction prevents two label collisions. An irregular rhythm is classified as 'HIGH reliability' when both modalities agree, whereas regular-looking PPG is classified as 'LOW reliability' when the ECG indicates hidden irregularity. Other relations are excluded from clear-label endpoints. The low-FPR analysis restricts further positives to hidden irregularity and negatives to both regular records, which is a narrower target than global LOW/HIGH AUROC. Full partition and case tables are provided in the Supplementary Material.

\paragraph{Hierarchical ECG teacher.}
The teacher processes eight contiguous 30-s ECG windows with a shared encoder, gated multi-view fusion, and a temporal Transformer, to produce both window-level and record-level three-class predictions. Averaged over three random seeds, it achieves a record-level rhythm  AUROC is $0.9912\pm0.0007$.
Architectural details and auxiliary training objectives are provided in the Supplementary Material.

\paragraph{PPG-only student.}
The student is initialized from SIGMA-PPG \cite{guo2026sigma}, pools 120 PPG patches into eight windows aligned with the teacher, and predicts the probability of low reliability ($p_{\mathrm{low}}$). B2 optimizes reliability BCE; the proof-of-concept K2 additionally matches aligned teacher outputs:
\begin{align}
\mathcal{L}_{K2}&=\mathcal{L}_{\mathrm{rel}}+0.15\cdot(
\mathcal{L}_{\mathrm{winKD}}+\mathcal{L}_{\mathrm{globalKD}}),\\
\mathcal{L}_{\mathrm{KD}}&=T^2\mathrm{KL}(p_T^{(T)}\|p_S^{(T)}),\\
p_\bullet^{(T)}&=\mathrm{softmax}(z_\bullet/T),\qquad T=2.
\end{align}
Here $z_T,z_S\in\mathbb{R}^{3}$ are teacher and student logits aligned by window  index. Window-level Knoledge Distillation (KD) averages the Kullback–Leibler divergence (KL) over the eight aligned windows, whereas global KD is applied once per record. K2-global-shuffled globally permutes the teacher targets while preserving architecture, marginal targets, losses, and optimization. B2, K2, and K2-global-shuffled share the same training schedule, optimization budget, data order, and random seeds (42, 123, and 2026). $T=2$ and the $0.15/0.15$ weights were selected using the validation set and frozen before access to the test set. Detailed architecture, index-alignment audit, hyperparameters, and the 17-arm ablation study are provided in the Supplementary Material

\paragraph{Evidence interface.}
The student output qualifies the use of the PPG-derived rhythm rather than
revealing an ECG diagnosis. The fixed language-facing block contains a
HIGH/LOW label, $p_{\mathrm{low}}$ rounded to four decimals, and an explanation
that the score estimates current-segment surrogate reliability. LOW denotes
$p_{\mathrm{low}}\geq0.5$ and HIGH otherwise; the frozen interface contains no
label--score mismatches. No ECG waveform, rhythm value, diagnosis, teacher
logit, or embedding is shown to the downstream LLM. Matched and shuffled
conditions preserve the case, baseline interaction, arm, seed, timing, and
template while changing only the patient-disjoint evidence source. The exact
block and complete prompts are provided in the Supplementary Material.

\paragraph{Worked metric example.}
In one D1 case, the PPG-derived rhythm is regular while the offline ECG
reference is irregular. The no-evidence verdict is
\texttt{TRUSTWORTHY}, so the case enters $\mathcal F_1$. Matched LOW evidence
changes the verdict to \texttt{SUSPECT}, giving
$\mathbf 1\{d_i(m)=y_i\}=1$, whereas shuffled HIGH evidence leaves the error
unchanged, giving $\mathbf 1\{d_i(s)=y_i\}=0$. The case therefore contributes
one to ESRM. D2 is evaluated analogously after severe history induces an error;
UHR instead uses the separate baseline-eligible set $\mathcal H$. Complete D1
and D2 examples are provided in the Supplementary Material.

\section{Results}

\begin{figure*}[!t]
\centering
\includegraphics[width=\textwidth]{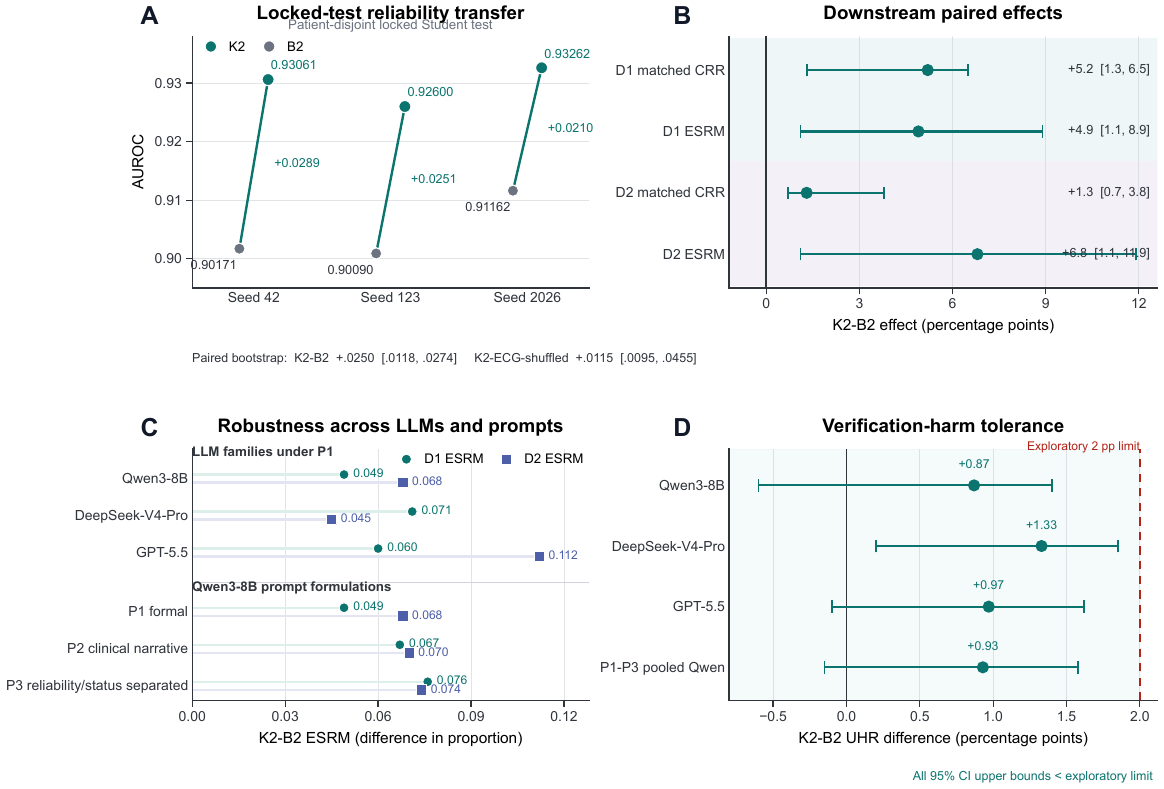}
\caption{DFOT results in the PPG--ECG case study. \textbf{(A)} Locked-test upstream reliability transfer. \textbf{(B)} Validation-stage conditional repair and evidence-package specificity. \textbf{(C)} Robustness across LLM families under P1 and across prompt formulations under Qwen3-8B. \textbf{(D)} Interface verification harm. The independent E3 replication is reported in Table~\ref{tab:dfot}.}
\label{fig:results}
\end{figure*}

\paragraph{DFOT induction.}
On validation, D1 COTR is 1.000 (300 records; 37 patients), and D2 CIR is 0.942 (292/310 neutral-correct records). E3 reproduces both manipulations: all 300 D1 records from 44 patients are over-trusted, while severe history induces an error in all 297 neutral-correct D2 records from 97 patients, giving CIR${}=1.000$. These controlled manipulation checks define the frozen error sets and are not prevalence estimates.

\begin{table*}[t]
\centering
\small
\setlength{\tabcolsep}{3.6pt}
\begin{tabular}{lrrrrl}
\toprule
Endpoint & Validation $\Delta$ [95\% CI] & E3 B2 & E3 K2 & E3 $\Delta$ [95\% CI] & Criterion \\
\midrule
D1 matched CRR & +5.2 [1.3,6.5] & 40.11 & 44.83 & +4.72 [1.67,7.28] & Met \\
D1 ESRM & +4.9 [1.1,8.9] & 21.67 & 26.17 & +4.50 [0.98,8.18] & Met \\
D2 matched CRR & +1.3 [0.7,3.8] & 94.39 & 96.21 & +1.82 [1.51,2.91] & Met \\
D2 ESRM & +6.8 [1.1,11.9] & 15.82 & 22.51 & +6.69 [2.60,7.19] & Met \\
UHR$^*$ & +0.87 [$-0.60$,1.40] & 2.89 & 3.56 & +0.67 [$-0.40$,1.70] & Met \\
\bottomrule
\end{tabular}
\caption{Controlled-validation effects and protocol-locked E3 replication. E3 B2/K2 columns are absolute percentages; deltas and CIs are percentage points. All efficacy CI lower bounds exceed zero. $^*$The UHR criterion is an exploratory interface-engineering limit (upper CI $<2$ pp), not clinical non-inferiority.}
\label{tab:dfot}
\end{table*}

\paragraph{Conditional repair and specificity.}
On controlled validation, all four K2--B2 efficacy intervals exclude zero. On E3, D1 matched CRR/ESRM increase from 40.11/21.67\% with B2 to 44.83/26.17\% with K2, corresponding to gains of 4.72 and 4.50 pp. Thus, 55.17\% of the deliberately induced D1 errors remain unrepaired, leaving substantial mitigation headroom. D2 matched CRR/ESRM increase from 94.39/15.82\% to 96.21/22.51\%, corresponding to gains of 1.82 and 6.69 pp. All four paired confidence intervals exclude zero, and every Student seed preserves the direction of effect.

\paragraph{Robustness across models and prompts.}
Under P1, the K2--B2 efficacy contrasts remain positive for Qwen3-8B, DeepSeek-V4-Pro, and GPT-5.5, with no directional reversal. Within Qwen3-8B, the effect also remains positive across P1--P3; pooled D1 and D2 ESRM gains are $+0.064$ [0.029, 0.097] and $+0.071$ [0.035, 0.108], respectively. This is not a complete LLM-by-prompt factorial evaluation.

\paragraph{The proof-of-concept supplies a discriminative baseline.}
The 17-arm development study selects K2 at AUROC of $.9042$ versus B2 at $.8927$, while within-case shuffled-target K5 remains close to $.9003$, suggesting that both privileged regularization and correct teacher--student correspondence contribute. On the independently locked upstream test, K2 exceeds B2 by $.0250$ AUROC [$.0118,.0274$] and globally shuffled ECG by $.0115$ [$.0095,.0455$]. The framework therefore distinguishes between evidence generators before and after the language interface. However, extreme low-FPR effects remain unresolved, meaning this discriminative baseline is a starting point rather than an endpoint for DFOT mitigation.

\paragraph{Negative controls sharpen the research target.}

A tail-aware objective improves the development tail mean by 3.06 percentage points (95\% CI 1.28--5.56) over K2 and by 2.59 points (95\% CI 0.85--4.53) over hidden-weighted BCE, but it does not demonstrate a reliable advantage over a patient-deranged shuffled-tail control: (+1.31) points (95\% CI (-1.70) to (+4.50)); pAUC difference (0.0057) (95\% CI (-0.0105) to (0.0212)). This control preserves the number of positive assignments (346) and all training mechanics while retaining the true hidden label for only 10.69\% of examples. We therefore do not pursue further tail-specific optimization. Instead, we conclude that the additional gain cannot be attributed to label-specific privileged information, highlighting the importance of negative controls in distinguishing genuine correspondence effects from generic optimization benefits.

\section{Discussion}

\paragraph{DFOT as a research target.}
The principal contribution of this work is a problem definition and measurement framework. Modular AI systems are commonly evaluated through either upstream prediction performance or downstream task accuracy, leaving the interface between them largely unexamined. An upstream estimate may be accurate on average yet unreliable for a particular instance, while a downstream model may reason coherently but assign that estimate more authority than its provenance supports. DFOT names this failure and separates its induction, repair, instance specificity, and interface cost into testable estimands. The matched--shuffled intervention further tests whether revision depends on evidence corresponding to the current instance rather than on the mere presence of reliability language.

\paragraph{What the case study establishes.}
The PPG--ECG case study shows that the framework is non-degenerate and discriminative. Both challenges reproducibly induce their intended failures, the metric chain distinguishes B2 from K2, and the comparison replicates on records from the patient-disjoint locked-test partition. The contrasts also persist across the tested LLM families under P1 and across prompt formulations within Qwen3-8B.

D1 and D2 probe complementary forms of inappropriate reliance: accepting a misleading derived measurement and allowing out-of-scope historical context to override a valid current measurement. K2 improves D1 matched repair and ESRM, although substantial residual error leaves clear headroom for stronger methods. D2 matched repair is already near ceiling, making its larger ESRM improvement more informative than its smaller CRR gain. K2 more strongly separates matched current-case evidence from a shuffled donor despite limited remaining repair headroom, illustrating why absolute CRR and ESRM should be interpreted jointly.

K2 should therefore be viewed as a reference baseline for the newly defined task rather than as a definitive solution. ECG-referenced distillation provides one convenient deployment-time PPG-only reliability signal with which to exercise the framework. The ablations suggest that regularization and category-level structure explain part of its benefit, while the locked global-shuffled comparison supports an additional contribution from correct teacher--student correspondence. Future DFOT methods need not use the same teacher, architecture, loss, or evidence representation.

\paragraph{Why a metric vector is necessary.}
Upstream transfer, downstream repair, evidence specificity, and interface cost capture different properties. Improved AUROC establishes better reliability ranking, not downstream use. Increased CRR establishes repair, but without ESRM it cannot distinguish matched-case evidence use from generic caution. UHR separately quantifies unnecessary verification on measurements that were already usable at baseline.

The unresolved low-FPR behavior reinforces this separation: global ranking can improve without improving the stringent operating region required by a particular workflow. DFOT mitigation is therefore multi-objective, involving failure coverage, repair, matched-case specificity, stringent-region behavior, and intervention cost. The exploratory 2-pp UHR limit is an interface-engineering constraint within this vector, not a universal clinical margin.

\paragraph{Transfer beyond physiological sensing.}
Although the empirical case study uses synchronized PPG and ECG, DFOT can arise whenever an upstream model produces an instance-dependent derived measurement that is consumed by a downstream reasoner. A new instantiation requires a derived measurement, an independently defined criterion for appropriate reliance, a controlled challenge exposing inappropriate reliance, and a matched--shuffled or equivalent counterfactual control that breaks instance correspondence while preserving the interface. The reference may come from a higher-fidelity modality, repeated measurements, expert adjudication, or another independently specified source. What transfers across domains is the problem definition, metric chain, and intervention logic; the particular reliability generator and numerical effects remain application-specific.

\paragraph{Research agenda.}
Future work can compare calibrated uncertainty heads, ensembles, conformal or selective predictors, retrieval-based verification, and stronger multimodal students within the same frozen downstream protocol. The purpose of the present study is not to declare DFOT solved by one ECG--PPG method, but to make a previously diffuse interface failure visible, measurable, and comparable. The formal definition, controlled challenges, linked estimands, and reference baseline establish a common target against which stronger mitigation methods can be developed.
\section{Ethical Statement}

This study uses retrospective, de-identified MIMIC waveform data which was accessed in accordance with the relevant credentials and data-use requirements. The study doesn't involve any prospective recruitment or clinical intervention. The proposed framework evaluates controlled LLM behavior and is not intended for use in direct clinical decision-making.

\section{Conclusion}

We introduce and formalize derived-feature over-trust, provide a five-estimand metric chain, and use matched-versus-shuffled interventions to distinguish matched-case repair from generic caution. A protocol-locked PPG--ECG case study demonstrates that the framework is reproducible and discriminative across tested patients, LLM families, and prompt formulations. The proof-of-concept K2 baseline improves reliability ranking, downstream repair, and evidence-package specificity while leaving clear headroom for stronger mitigation methods. Although demonstrated with physiological signals, the DFOT definition and evaluation chain apply more broadly wherever downstream models reason over instance-dependent derived measurements. DFOT is therefore a concrete evaluation target and an open research agenda for more reliable measurement-to-LLM pipelines.

\bibliographystyle{plainnat}
\bibliography{references}

\clearpage
\onecolumn
\appendix
\setcounter{secnumdepth}{1}
\setcounter{topnumber}{5}
\setcounter{bottomnumber}{5}
\setcounter{totalnumber}{10}
\renewcommand{\topfraction}{0.95}
\renewcommand{\bottomfraction}{0.95}
\renewcommand{\textfraction}{0.05}
\renewcommand{\floatpagefraction}{0.85}
\section*{Supplementary Material}
\section{Scope and Reading Guide}

This supplement documents the experiment ledger, complete arm definitions, protocol lock, negative controls, and analysis boundaries. The paper's contribution hierarchy is: (1) naming and formalizing DFOT; (2) defining its five-estimand metric chain; (3) introducing matched-versus-shuffled evidence as an instance-specificity intervention; and (4) validating the framework through a PPG--ECG case study. ECG-to-PPG distillation is the focused proof-of-concept baseline used in that fourth step, not the general definition of DFOT.

For auditability, the six result families below are listed in experimental chronology rather than order of conceptual importance:
\begin{enumerate}
    \item \textbf{Student validation}: arm and threshold selection, including the 17-arm ablation and exact original-recipe reproduction.
    \item \textbf{Locked upstream test}: a one-shot evaluation of frozen B2, K2, and shuffled-ECG checkpoints. No downstream LLM calls had been made on the test split at this stage.
    \item \textbf{Controlled DFOT validation (E0)}: characterization, mitigation, evidence-package specificity, and interface verification-harm endpoints under Qwen3-8B/P1.
    \item \textbf{Cross-LLM robustness (E1)}: the frozen comparisons under Qwen3-8B, DeepSeek-V4-Pro, and GPT-5.5.
    \item \textbf{Prompt robustness (E2)}: Qwen3-8B under three semantically equivalent prompt formulations.
    \item \textbf{Downstream test replication (E3)}: a separately protocol-locked Qwen3-8B/P1 evaluation on the previously downstream-unqueried patient-disjoint test partition.
\end{enumerate}

\section{Data, Splits, and Case Construction}

\subsection{Patient-disjoint cohort}

The source is the MIMIC-III Matched Waveform Database v1.0. A channel-level scan identified 1,297 candidate patients with simultaneous PLETH and Lead-II recordings. Requiring a complete 240-s paired interval, at most 20\% missing samples in either raw channel, and valid ECG and PPG beat-interval extraction yielded the frozen 50,000-record cohort from 1,275 patients. Missing samples in retained intervals were linearly interpolated; the frozen 50k table and patient manifest contain no missing identifiers or required endpoint fields. PPG is resampled to 50 Hz and ECG to 125 Hz. No MIMIC clinical table enters the frozen waveform builder, so cohort age, sex, and race summaries are unavailable. All splitting is performed at patient level.

\begin{center}
\centering
\setlength{\tabcolsep}{3pt}
\begin{tabular}{lrrl}
\toprule
Split & Records & Patients & Role \\
\midrule
Train & 36,115 & 925 & Training \\
Validation & 6,464 & 163 & Selection \\
Locked test & 7,421 & 187 & Confirmation \\
Total & 50,000 & 1,275 & Full cohort \\
\bottomrule
\end{tabular}
\captionof{table}{Frozen cohort partitions after the 1,297-patient channel-level source screen. Patient overlap is zero for every pair of splits.}
\label{tab:splits}
\end{center}

The primary reliability label is defined only on clear ECG/PPG rhythm relations. Rows outside the four clear cases remain in the protocol data but are excluded from the clear-label reliability endpoints.

\begin{center}
\centering
\setlength{\tabcolsep}{3pt}
\begin{tabular}{llll}
\toprule
Case & PPG & ECG ref. & Reliability \\
\midrule
Both regular & Regular & Regular & HIGH \\
Hidden irreg. & Regular & Irregular & LOW \\
Both irregular & Irregular & Irregular & HIGH \\
False alarm & Irregular & Regular & LOW \\
\bottomrule
\end{tabular}
\captionof{table}{Clear rhythm cases. Reliability denotes agreement with the ECG-referenced rhythm, not clinical normality.}
\label{tab:cases}
\end{center}

On the locked test, 3,733 rows satisfy the clear-case definition: 897 LOW and 2,836 HIGH. The low-FPR contrast specifically uses 401 hidden-irregularity and 2,151 both-regular rows. The distinction matters because a model can rank the full LOW/HIGH task well while failing to isolate hidden irregularity at stringent regular-case FPR.

\section{Hierarchical Teacher and Student}

\subsection{ECG teacher}

The teacher receives 30,000 ECG samples and partitions them into eight contiguous 30-s windows. Each window is represented by overlapping 0--25 s and 5--30 s views. A shared encoder processes both views; gated fusion produces one local representation. A two-layer temporal Transformer aggregates the eight representations into a global state. The outputs are:
\begin{itemize}
    \item eight 256-dimensional window embeddings and one global embedding;
    \item window and global rhythm logits;
    \item heart rate, RMSSD, and SDNN estimates.
\end{itemize}

The teacher is task-specific and trained from scratch. A public pretraining initialization did not improve this task and was not used for the primary teacher. Three-seed mean global rhythm AUROC is $0.9912\pm0.0007$, HR MAE is $1.241\pm0.004$ bpm, RMSSD correlation is $0.9868\pm0.0011$, and SDNN correlation is $0.9831\pm0.0014$. These values qualify the teacher as a privileged reference model; they do not demonstrate transfer.

\subsection{Exact original-recipe validation}
\begin{table}[!htbp]
\centering
\setlength{\tabcolsep}{4.2pt}
\begin{tabular}{lrrrrrrr}
\toprule
Arm & AUROC & AUPRC & Brier & ECE15 & TPR@1\% & TPR@2\% & TPR@5\% \\
\midrule
B2 & .8926 & .7466 & .1018 & .0689 & 19.05\% & 25.87\% & 36.80\% \\
K2-global-shuffled & .8909 & .7509 & .1000 & .0710 & 17.53\% & 22.62\% & 37.34\% \\
K2 matched & .9018 & .7642 & .0984 & .0577 & 17.86\% & 23.81\% & 36.15\% \\
\bottomrule
\end{tabular}
\caption{Exact original-recipe validation. K2 improves global ranking and calibration but not hidden-irregularity sensitivity in the extreme low-FPR region.}
\label{tab:validation}
\end{table}

\subsection{PPG student and objectives}

The PPG input has 12,000 samples. Patch size 100 yields 120 tokens from the 18-block SIGMA-PPG encoder. Eight groups of 15 tokens are attention-pooled to align with teacher windows. A two-layer, 256-dimensional, eight-head temporal Transformer produces window and global states. The primary output is $p_{\mathrm{low}}$.

For a clear-case reliability target $y\in\{0,1\}$, the base objective is weighted BCE:
\begin{equation}
 \mathcal{L}_{\mathrm{rel}}=\mathrm{BCEWithLogits}(r,y; w_+=N_{\mathrm{HIGH}}/N_{\mathrm{LOW}}).
\end{equation}
For three-dimensional teacher and student rhythm logits
$z_T,z_S\in\mathbb{R}^{3}$ and $T=2$,
\begin{equation}
 \mathcal{L}_{\mathrm{KD}}=T^2\mathrm{KL}\left(\mathrm{softmax}(z_T/T)\Vert\mathrm{softmax}(z_S/T)\right).
\end{equation}
Window KD is averaged across the eight windows; global KD is computed once per record. The scalar reliability logit $r$ is produced by a separate head and is supervised by $\mathcal{L}_{\mathrm{rel}}$ rather than by the logit losses. The K2 objective is $\mathcal{L}_{\mathrm{rel}}+0.15\mathcal{L}_{\mathrm{winKD}}+0.15\mathcal{L}_{\mathrm{globalKD}}$. The audit confirms output shapes $[N,8,3]$ and $[N,3]$ and that teacher/student indices are aligned for KL matching, but it does not recover the three class names/index order, the exact teacher-label generation rule, or whether the window/global classifier parameters are shared. A common permutation of aligned teacher and student indices leaves the KL objective unchanged; this mathematical invariance does not recover class-specific clinical semantics or complete source-level reproducibility. Pending recovery of the training sources, we therefore interpret K2 as a frozen aligned three-way auxiliary-logit recipe and make no index-specific mechanism claim.

The temperature $T=2$ and the equal window/global coefficients $0.15/0.15$ were chosen from a limited candidate set during validation-stage development and frozen before locked-test access. The locked test was not used for hyperparameter selection. Because the search was limited rather than exhaustive, the reported findings establish the behavior of this frozen recipe but do not establish hyperparameter optimality or broad insensitivity.

\section{Proof-of-Concept Student Ablations}

\subsection{Training schedule}

The exact original recipe uses maximum 15 epochs, patience 4, batch size 32, weight decay 0.01, and gradient clipping at 1.0. The head and backbone learning rates are $10^{-4}$ and $10^{-5}$. Epochs 1--2 train the head, epochs 3--5 partially unfreeze the final six backbone blocks, and the remaining epochs permit full adaptation. Seeds are 42, 123, and 2026. The locked test uses nine frozen checkpoints: three seeds for B2, K2, and K2-global-shuffled.

\begin{table}[!htbp]
\centering
\small
\setlength{\tabcolsep}{5pt}
\begin{tabular}{llllrr}
\toprule
Arm & Backbone & Reliability supervision & Privileged target & AUROC & AUPRC \\
\midrule
B1 & Frozen SIGMA & BCE & None & $.8887\pm.0079$ & -- \\
B2 & Fine-tuned SIGMA & BCE & None & $.8927\pm.0012$ & $.7512\pm.0041$ \\
K1 & Fine-tuned SIGMA & BCE & HR/RMSSD/SDNN & $.8949\pm.0032$ & -- \\
K2 & Fine-tuned SIGMA & BCE & Window + global rhythm logits & $\mathbf{.9042\pm.0046}$ & $\mathbf{.7639\pm.0087}$ \\
K3 & Fine-tuned SIGMA & BCE & Window + global embeddings & $.8960\pm.0033$ & -- \\
H1 & Fine-tuned SIGMA & BCE & Global rhythm logits & $.8995\pm.0008$ & -- \\
H2 & Fine-tuned SIGMA & BCE & Window rhythm logits & $.9015\pm.0033$ & -- \\
K5 & Fine-tuned SIGMA & BCE & Within-case shuffled full target & $.9003\pm.0053$ & -- \\
K4 & Fine-tuned SIGMA & BCE & Logits + embeddings + metrics & $.9018\pm.0031$ & $.7522\pm.0177$ \\
S4 & Scratch & BCE & Full hierarchical target & $.8755\pm.0069$ & $.7190\pm.0110$ \\
\bottomrule
\end{tabular}
\caption{Historical 17-arm validation ablation (principal arms shown). K2 is the selected student. ``Full'' supervision is not uniformly better, consistent with optimization conflict among heterogeneous teacher targets.}
\label{tab:ablation}
\end{table}

The ablation supports three limited interpretations. First, SIGMA initialization matters because K4 exceeds its scratch counterpart S4. Second, both temporal levels contribute descriptively: K2 exceeds H1 and H2. Third, the aligned-logit recipe is more effective than explicit metrics or latent imitation in this development screen. K5 preserves rhythm case while shuffling full targets within case and remains close to K2 ($.9003$ versus $.9042$ AUROC). Thus the ablation cannot attribute the full K2 gain to instance-matched privileged semantics; category-level structure, auxiliary-task regularization, or optimization effects plausibly contribute. The locked K2-versus-global-shuffled comparison tests whether an additional instance-correspondence component remains, but it does not decompose these mechanisms.

K2--B2 validation AUROC and AUPRC differences are $+0.0093$ and $+0.0176$. K2--shuffled AUROC is $+0.0110$. At threshold 0.65, K2--B2 hidden LOW is $-2.49$ points (95\% CI $-9.63$--$5.34$), and operational H-mean is $-3.13$ points ($-12.00$--$6.36$). The frozen validation decision is therefore ranking-only.

\section{Locked-Test Protocol and Results}
\begin{table}[!htbp]
\centering
\setlength{\tabcolsep}{5pt}
\begin{tabular}{rrrr}
\toprule
Seed & B2 AUROC & K2 AUROC & K2--B2 \\
\midrule
42 & .90171 & .93061 & +.02890 \\
123 & .90090 & .92600 & +.02510 \\
2026 & .91162 & .93262 & +.02100 \\
\bottomrule
\end{tabular}
\caption{Frozen seed-level AUROC on the patient-disjoint locked Student test. The K2--B2 difference is positive for every seed.}
\label{tab:test-seeds}
\end{table}

The protocol fixed data manifests, thresholds, model identifiers, checkpoint hashes, validation prediction hashes, comparison signs, and bootstrap settings before test access. The test was accessed once under lock hash
\begin{center}
\texttt{e81efb2019fbbdc326aa7d34da195bec\allowbreak62ea25012386f01a76b178de826ed427}.
\end{center}
All nine checkpoints completed. The test was not used for retraining, checkpoint choice, threshold selection, prompt design, or downstream LLM evaluation.

\begin{table}[!htbp]
\centering
\setlength{\tabcolsep}{4pt}
\begin{tabular}{llrrr}
\toprule
Comparison & Metric & Mean difference & 95\% CI & Interpretation \\
\midrule
K2--B2 & AUROC & +.0250 & [.0118,.0274] & Confirmed global ranking \\
K2--ECG-shuffled & AUROC & +.0115 & [.0095,.0455] & Confirmed ECG specificity \\
K2--B2 & standardized pAUC@5\% FPR & +.0023 & [-.0409,.0446] & Not confirmed \\
K2--B2 & TPR@1\% FPR & -.0008 & [-.1763,.1620] & Not confirmed \\
K2--B2 & TPR@2\% FPR & +.0037 & [-.1642,.1359] & Not confirmed \\
K2--B2 & TPR@5\% FPR & +.0082 & [-.0801,.0957] & Not confirmed \\
\bottomrule
\end{tabular}
\caption{Frozen one-shot locked-test paired hierarchical-bootstrap effects. Matched K2 improves global ranking over both B2 and ECG-shuffled control; the low-FPR intervals rule out a claim that K2 improves all safety-relevant operating points.}
\label{tab:test}
\end{table}

\section{DFOT Tasks and Metrics}

\subsection{Task separation}

Characterization is run before mitigation and without reliability instructions. This is necessary: a baseline prompt that already tells the LLM to verify whenever evidence is missing destroys the over-trusted subset and makes CRR undefined. D1 and D2 use distinct prompts because they test different mechanisms. Utility is a third task with separate outputs for measurement use and clinical status.

The exact frozen prompt strings and deterministic parsers are versioned protocol artifacts. No separate system prompt is used; task instructions are carried by the user messages. Their semantic skeleton is:
\begin{itemize}
    \item \textbf{D1 characterization}: current PPG-derived rhythm + decision question; no reliability evidence.
    \item \textbf{D2 characterization}: the identical current segment under neutral versus severe history; no reliability evidence.
    \item \textbf{Reactive mitigation}: exact previous answer + current matched or patient-disjoint shuffled reliability statement.
    \item \textbf{Utility}: current measurement + evidence; output MEASUREMENT\_USE and CLINICAL\_STATUS separately.
\end{itemize}
The confirmed reliability block is reproduced below. The submission artifact includes the exact P1--P3 task templates, model request records, parser definitions, case manifests, donor assignments, and their frozen hashes.

\subsection{Qwen checkpoint and inference runtime}

E0 and the Qwen portions of E1/E2 use a local five-shard bfloat16 Qwen3-8B checkpoint with \texttt{Qwen3ForCausalLM}, 36 layers, hidden size 4096, 32 attention heads, eight key-value heads, and maximum context length 40,960. It is served through the OpenAI-compatible vLLM 0.25.1 runtime. Formal requests explicitly set temperature to 0, maximum output length to 96 tokens, and thinking mode to disabled. Although the checkpoint's bundled generation configuration contains sampling defaults, these are not the formal request settings and do not alter greedy temperature-zero decoding.

The retained SHA-256 values are \texttt{f7c4...7c30} for \texttt{config.json}, \texttt{989f...f654} for the frozen P1 prompt code, \texttt{7177...17fb} for the formal configuration, and \texttt{be0f...e51c} for the deterministic scorer. The local checkpoint lacks a Hugging Face snapshot reference, so no exact Hub revision is claimed. The \texttt{transformers\_version=4.51.0} field belongs to the saved model configuration and is not presented as the executing runtime version.

\subsection{Reliability-evidence interface}

The downstream LLM receives a fixed textual rendering of the deployment-time PPG-only reliability student. The label is generated directly from the displayed score: LOW when $p_{\mathrm{low}}\geq0.5$ and HIGH otherwise. Scores are rounded to four decimal places, and the frozen downstream export has zero label--score mismatches. Student-validation operating thresholds are not used for this language-facing rendering. The exact evidence block is:

\begin{samepage}\begin{quote}\small\ttfamily
Additional current-segment reliability evidence is now available:\\
Current-segment rhythm reliability evidence: HIGH/LOW.\\
Estimated probability of LOW rhythm reliability: 0.xxxx.\\
This evidence estimates whether the PPG-derived rhythm feature is a reliable surrogate for ECG-referenced rhythm in this current segment.\\
HIGH/LOW describes surrogate reliability only; the ECG rhythm value itself is not shown.
\end{quote}\end{samepage}

In an instantiated prompt, HIGH/LOW and 0.xxxx are replaced by the corresponding entries in Table~\ref{tab:evidence-examples}; all other wording is identical. Audit identifiers, student arm, seed, and donor identity are not rendered. The LLM receives the case text, PPG-derived heart rate, categorical PPG-derived rhythm, and this PPG-only evidence. It receives no ECG waveform, ECG rhythm value, ECG-derived diagnosis or label, teacher logit, teacher embedding, or teacher window/global representation.

\begin{table}[!htbp]
\centering
\small
\setlength{\tabcolsep}{3.5pt}
\begin{tabular}{llrrl}
\toprule
Case & Evidence & Label & $p_{\mathrm{low}}$ & Verdict \\
\midrule
D1 & matched  & LOW  & .9961 & SUSPECT \\
D1 & shuffled & HIGH & .0003 & TRUSTWORTHY \\
D2 & matched  & HIGH & .1465 & TRUSTWORTHY \\
D2 & shuffled & HIGH & .3164 & SUSPECT \\
\bottomrule
\end{tabular}
\caption{Exact substitutions in the fixed evidence template and resulting parsed verdicts for the worked examples.}
\label{tab:evidence-examples}
\end{table}

\paragraph{D1 matched and shuffled example.}
For audit case \texttt{val\_hidden\_irregularity\_1155} (K2, seed 123, reactive), the language-facing PPG rhythm is regular while the offline ECG reference is irregular. The no-evidence response is:
\begin{quote}\small\ttfamily
VERDICT: TRUSTWORTHY\\
ACTION: The PPG-derived rhythm assessment is reliable for routine use in this segment.
\end{quote}
The instance-matched LOW/.9961 evidence changes the response to:
\begin{quote}\small\ttfamily
VERDICT: SUSPECT\\
ACTION: The PPG-derived rhythm assessment has low reliability and should be interpreted with caution.
\end{quote}
Thus $R_i(\mathrm{matched})=1$. For the patient-disjoint shuffled donor, the same case, frozen baseline, arm, seed, timing, and prompt are retained. HIGH/.0003 evidence leaves the response unchanged:
\begin{quote}\small\ttfamily
VERDICT: TRUSTWORTHY\\
ACTION: The PPG-derived rhythm assessment is reliable for routine use in this segment.
\end{quote}
Hence $R_i(\mathrm{shuffled})=0$ and this case contributes $1$ to ESRM.

\paragraph{D2 matched and shuffled example.}
For audit case \texttt{val\_both\_regular\_49505} (K2, seed 2026, reactive), the neutral-history no-evidence response correctly returns TRUSTWORTHY. Severe history changes the no-evidence response to:
\begin{quote}\small\ttfamily
VERDICT: SUSPECT\\
ACTION: Cross-check with clinical context and consider potential PPG artifact or undetected arrhythmia.
\end{quote}
The case therefore enters the frozen D2 induced set. Matched HIGH/.1465 evidence produces the revised response:
\begin{quote}\small\ttfamily
VERDICT: TRUSTWORTHY\\
ACTION: Current segment rhythm assessment is reliable based on high surrogate reliability evidence.
\end{quote}
Patient-disjoint shuffled HIGH/.3164 evidence instead leaves the response incorrect:
\begin{quote}\small\ttfamily
VERDICT: SUSPECT\\
ACTION: Consider potential undetected arrhythmia despite high rhythm reliability evidence.
\end{quote}
Hence $R_i(\mathrm{matched})=1$, $R_i(\mathrm{shuffled})=0$, and the case contributes $1$ to ESRM. This example shows different revisions when both labels are HIGH but displayed scores differ; it does not by itself establish the continuous score as a general causal mechanism.

\paragraph{Utility interface.}
Utility responses use separate fields, for example:
\begin{quote}\small\ttfamily
MEASUREMENT\_USE: USE\_AS\_IS\\
CLINICAL\_STATUS: REGULAR
\end{quote}
or
\begin{quote}\small\ttfamily
MEASUREMENT\_USE: VERIFY\_FIRST\\
CLINICAL\_STATUS: UNDETERMINED
\end{quote}
The same evidence block is used, but the denominator is a separate record-disjoint HIGH-reliability utility set whose baseline measurement use is USE\_AS\_IS. UHR counts only evidence-induced VERIFY\_FIRST; the separately parsed clinical-status field does not define utility harm. UHR is therefore an interface-specific unnecessary-verification endpoint, not a comprehensive measure of clinical utility, delayed care, escalation, or patient outcome.

\subsection{Metric denominators}

Let $\mathcal{C}$ denote the reference-confirmed D1 conflict cases and
$\mathcal{N}$ the D2 cases whose no-evidence, neutral-context decision is
correct. Let $b_i$ denote the parsed downstream decision in the corresponding
no-evidence challenge condition, $d_i^{\mathrm{severe}}$ the parsed decision
in the no-evidence severe-history condition, and $d_i(e)$ the parsed
downstream decision under evidence condition $e$. Let $u_i(e)$ denote the
parsed \texttt{MEASUREMENT\_USE} field under evidence condition $e$, and let
$y_i$ denote the parser-defined target decision
(\textsc{Suspect} for D1 and \textsc{Trustworthy} for D2).

Let
$z_i\in\{\text{\textsc{High}},\text{\textsc{Low}}\}$ denote the
offline reference-defined reliability class, which is distinct from the
deployment-time reliability estimate $R_i$. The baseline-eligible utility set is
\begin{equation}
\mathcal{H}
=
\left\{
i:z_i=\text{\textsc{High}}
\ \land\
u_i(\varnothing)=\text{\textsc{Use-As-Is}}
\right\}.
\end{equation}

The challenge-specific frozen error sets are
\[
\mathcal F_1
=
\{i\in\mathcal C:b_i\neq y_i\},
\qquad
\mathcal F_2
=
\{i\in\mathcal N:d_i^{\mathrm{severe}}\neq y_i\}.
\]
We use $\mathcal F=\mathcal F_1$ for D1 and
$\mathcal F=\mathcal F_2$ for D2, and abbreviate matched and shuffled
evidence by $m$ and $s$, respectively. The five estimands are
\begin{align}
\mathrm{COTR}
&=
\frac{\#\{i\in\mathcal C:\text{baseline accepts }D_i\}}
     {|\mathcal C|},\\
\mathrm{CIR}
&=
\frac{\#\{i\in\mathcal N:d_i^{\mathrm{severe}}\neq y_i\}}
     {|\mathcal N|},\\
\mathrm{CRR}(e)
&=
\frac{\#\{i\in\mathcal F:d_i(e)=y_i\}}
     {|\mathcal F|},\\
\mathrm{UHR}
&=
\frac{\#\{i\in\mathcal H:
u_i(m)=\text{\textsc{Verify-First}}\}}
     {|\mathcal H|}.
\end{align}
ESRM is defined as
\[
\mathrm{ESRM}
=
\mathrm{CRR}(m)-\mathrm{CRR}(s).
\]

The randomized source substitution preserves the case, frozen baseline, student arm, seed, timing, and evidence template while replacing only the patient-disjoint evidence source. Consequently, the rendered HIGH/LOW label
and numerical score may change jointly. ESRM therefore measures the benefit of the complete matched evidence package relative to the shuffled package.
The K2--B2 ESRM contrast quantifies the additional evidence-package specificity associated with teacher supervision beyond the generic effect of providing reliability evidence. It does not separately identify the contributions of the categorical label, numerical score, thresholding, or template wording, and no factorial decomposition of the evidence interface is claimed.

For a concrete D1 case, a no-evidence \texttt{TRUSTWORTHY} verdict on a reference-confirmed conflict contributes one event to COTR and places the
instance in $\mathcal F_1$. If matched evidence restores the target decision
(\texttt{SUSPECT}) while shuffled evidence leaves the baseline decision unchanged, then the case contributes
\[
\mathbf{1}\{d_i(m)=y_i\}
-
\mathbf{1}\{d_i(s)=y_i\}
=
1
\]
to the ESRM numerator. In D2, a neutral-correct case that becomes incorrect under severe history contributes one event to CIR and enters $\mathcal F_2$;
matched and shuffled evidence are then evaluated on that frozen induced case.
Individual ESRM contributions can therefore be $-1$, $0$, or $1$, and the reported ESRM averages these contributions over the frozen error set. UHR is computed on the independent utility set $\mathcal H$, not on either repair set.

\begin{table}[!htbp]
\centering
\setlength{\tabcolsep}{4.5pt}
\begin{tabular}{llrrrrr}
\toprule
Scenario & Arm & COTR/CIR & Matched CRR & Shuffled CRR & ESRM & UHR \\
\midrule
D1 & B2 & 1.000 & .298 & .171 & .127 & 3.00\% \\
D1 & K2 & 1.000 & .341 & .159 & .182 & 3.92\% \\
D1 & K4 & 1.000 & .323 & .200 & .123 & 4.44\% \\
D2 & B2 & .942 & .947 & .794 & .153 & 3.00\% \\
D2 & K2 & .942 & .961 & .747 & .214 & 3.92\% \\
D2 & K4 & .942 & .943 & .796 & .147 & 4.44\% \\
\bottomrule
\end{tabular}
\caption{Aggregate formal v3.1 reactive results. COTR and CIR are challenge-set manipulation checks and therefore do not vary by student arm; they are not prevalence estimates. Arm-level rates are descriptive aggregates; the prespecified paired hierarchical-bootstrap effects are reported in Table~\ref{tab:gates}.}
\label{tab:formal}
\end{table}

\section{DFOT Metric Closure on Controlled Validation}

D1 contains 300 hidden-conflict cases from 37 patients, all over-trusted at baseline. D2 has 310 neutral-correct eligible cases; severe history induces errors in 292 cases from 89 patients, for CIR 0.942 (95\% CI 0.877--0.994).

In D1, K2 increases matched CRR by 5.2 points and ESRM by 4.9 points over B2, supporting partial correct-repair and matched-package-specificity gains in current-measurement conflicts. The D2 matched CRR effect is small in absolute magnitude because B2 already repairs 94.7\% of the induced set. ESRM is the more diagnostic endpoint: K2 increases the matched--shuffled package gap by 6.8 points under a ceilinged repair rate. The UHR increase is $0.0087$ (95\% CI $-0.006$--$0.014$); its upper bound meets the prespecified exploratory 0.02 interface-engineering limit. This does not establish zero harm, broader utility, or clinical non-inferiority.

\begin{table}[!htbp]
\centering
\setlength{\tabcolsep}{3pt}
\begin{tabular}{lrrl}
\toprule
Prespecified endpoint & Mean & 95\% CI & Gate \\
\midrule
D1 K2--B2 CRR & +.052 & [.013,.065] & Pass \\
D1 K2--B2 ESRM & +.049 & [.011,.089] & Pass \\
D2 K2--B2 CRR & +.013 & [.007,.038] & Pass \\
D2 K2--B2 ESRM & +.068 & [.011,.119] & Pass \\
K2--B2 UHR & +.0087 & [-.006,.014] & Pass$^*$ \\
\bottomrule
\end{tabular}
\caption{Paired patient/seed hierarchical bootstrap. $^*$Upper CI below the prespecified exploratory 0.02 interface-engineering limit, not a clinical non-inferiority margin.}
\label{tab:gates}
\end{table}

\section{E1: Cross-LLM Robustness}

E1 reuses the frozen patients, cases, student seeds, matched/shuffled assignments, parser, and endpoints under P1 while changing only the LLM family. The three LLMs form a fixed robustness panel rather than a random sample from a population of models.

\begin{table}[!htbp]
\centering
\setlength{\tabcolsep}{5pt}
\begin{tabular}{lrrr}
\toprule
Endpoint & Qwen3-8B & DeepSeek-V4-Pro & GPT-5.5 \\
\midrule
D1 COTR & 1.000 & .940 & 1.000 \\
D2 CIR & .942 & .829 & .931 \\
D1 K2 matched CRR & .341 & .372 & .390 \\
D1 K2 ESRM & .182 & .151 & .248 \\
D1 K2--B2 ESRM & +.049 & +.071 & +.060 \\
D2 K2 matched CRR & .961 & .929 & .951 \\
D2 K2 ESRM & .214 & .276 & .314 \\
D2 K2--B2 ESRM & +.068 & +.045 & +.112 \\
K2--B2 UHR & +.87 pp & +1.33 pp & +.97 pp \\
Parse failures & 0 & 0 & 0 \\
\bottomrule
\end{tabular}
\caption{E1 descriptive cross-LLM results. DFOT remains inducible and K2--B2 ESRM remains positive in both scenarios for every tested model.}
\label{tab:e1-descriptive}
\end{table}

\begin{table}[!htbp]
\centering
\scriptsize
\setlength{\tabcolsep}{3.2pt}
\begin{tabular}{lrrrr}
\toprule
Endpoint & Qwen3-8B $\Delta$ [95\% CI] & DeepSeek-V4-Pro $\Delta$ [95\% CI] & GPT-5.5 $\Delta$ [95\% CI] & Equal-model pooled $\Delta$ [95\% CI] \\
\midrule
D1 matched CRR & +.052 [.013,.065] & +.064 [.021,.104] & +.072 [.028,.113] & +.063 [.033,.087] \\
D1 ESRM & +.049 [.011,.089] & +.071 [.018,.126] & +.060 [.016,.108] & +.060 [.030,.091] \\
D2 matched CRR & +.013 [.007,.038] & +.021 [.004,.046] & +.028 [.010,.051] & +.021 [.010,.036] \\
D2 ESRM & +.068 [.011,.119] & +.045 [.009,.087] & +.112 [.054,.171] & +.075 [.041,.111] \\
\bottomrule
\end{tabular}

\vspace{4pt}
\begin{tabular}{lrrl}
\toprule
Model & K2--B2 UHR & 95\% CI & Gate \\
\midrule
Qwen3-8B & +.87 pp & [-.60,+1.40] pp & Pass \\
DeepSeek-V4-Pro & +1.33 pp & [+.20,+1.85] pp & Pass \\
GPT-5.5 & +.97 pp & [-.10,+1.62] pp & Pass \\
\bottomrule
\end{tabular}
\caption{E1 paired hierarchical-bootstrap inference. Pooled efficacy effects are equal-weight averages over the three LLMs within synchronized patient/seed bootstrap replicates. Every efficacy CI lower bound exceeds zero, and every UHR CI upper bound is below the 2-pp tolerance.}
\label{tab:e1-inference}
\end{table}

The E1 claim is limited to the three tested LLMs. It supports cross-family robustness of the controlled effect, not invariance across all present or future LLMs.

\section{E2: Prompt Robustness}

E2 holds Qwen3-8B, medical facts, evidence content, cases, target task, and parser fixed. P1 is the formal template; P2 reorganizes the same content as a clinical narrative; P3 explicitly separates measurement reliability from current clinical status. Only wording and organization change.

\begin{table}[!htbp]
\centering
\setlength{\tabcolsep}{4pt}
\begin{tabular}{lp{2.65in}}
\toprule
Prompt & Frozen formulation constraint \\
\midrule
P1 & Current formal template \\
P2 & Clinically narrative expression of identical facts and task \\
P3 & Explicit separation of measurement reliability and current clinical status \\
\bottomrule
\end{tabular}
\caption{E2 prompt formulations. Medical content and target task are invariant.}
\label{tab:e2-prompts}
\end{table}

\begin{table}[!htbp]
\centering
\setlength{\tabcolsep}{5pt}
\begin{tabular}{lrrr}
\toprule
Endpoint & P1 & P2 & P3 \\
\midrule
D1 COTR & 1.000 & .930 & .880 \\
D2 CIR & .942 & .860 & .790 \\
D1 K2 matched CRR & .341 & .365 & .395 \\
D1 K2 ESRM & .182 & .189 & .211 \\
D1 K2--B2 ESRM & +.049 & +.067 & +.076 \\
D2 K2 matched CRR & .961 & .949 & .970 \\
D2 K2 ESRM & .214 & .217 & .254 \\
D2 K2--B2 ESRM & +.068 & +.070 & +.074 \\
K2--B2 UHR & +.87 pp & +.95 pp & +.95 pp \\
Parse failures & 0 & 0 & 0 \\
\bottomrule
\end{tabular}
\caption{E2 prompt-specific results. DFOT remains above the prespecified inducibility thresholds, and no efficacy endpoint reverses direction.}
\label{tab:e2-descriptive}
\end{table}

\begin{table}[!htbp]
\centering
\setlength{\tabcolsep}{3pt}
\begin{tabular}{lrrl}
\toprule
Pooled endpoint & $\Delta$ & 95\% CI & Gate \\
\midrule
D1 matched CRR & +.059 & [.025,.087] & Pass \\
D1 ESRM & +.064 & [.029,.097] & Pass \\
D2 matched CRR & +.024 & [.008,.040] & Pass \\
D2 ESRM & +.071 & [.035,.108] & Pass \\
K2--B2 UHR & +.93 pp & [-.15,+1.58] pp & Pass \\
\bottomrule
\end{tabular}
\caption{E2 equal-prompt pooled paired effects. All efficacy CI lower bounds exceed zero; the UHR CI upper bound is below 2 pp.}
\label{tab:e2-inference}
\end{table}

P3 has the largest descriptive ESRM values, but E2 does not prespecify or test P3 superiority over P1/P2. The supported claim is persistence across semantically equivalent formulations, not universal prompt invariance.

\section{E3: Protocol-Locked Downstream Test Replication}

\subsection{Scope and freeze}

E3 evaluates downstream transport to the MIMIC test partition, which is patient-disjoint from training and validation. The partition's upstream Student predictions had already been evaluated, but no downstream LLM response from these patients had been accessed. Before inference, case-selection seed 20260717 fixed the target records, donor rules, parser, bootstrap, four efficacy gates, and one exploratory interface criterion. The protocol lock records SHA-256 \texttt{c1b1...d735}; the combined-case manifest is \texttt{b629...bd77}. E3 uses the same Qwen3-8B/P1 interface, B2/K2 checkpoints, three Student seeds, reactive timing, and 0.5 evidence threshold as the controlled analysis.

\begin{table}[!htbp]
\centering
\setlength{\tabcolsep}{4pt}
\begin{tabular}{lrrr}
\toprule
Task & Selected records & Selected patients & Frozen analysis set \\
\midrule
D1 hidden irregularity & 300 & 44 & 300 / 44 patients \\
D2 both regular & 300 & 99 & 297 / 97 patients \\
Utility HIGH & 300 & 111 & 150 / 76 patients \\
\midrule
Record-disjoint union & 900 & 158 & --- \\
\bottomrule
\end{tabular}
\caption{E3 target sets. Tasks are record-disjoint but can contain different records from the same patient. D1/D2 frozen sets are baseline errors; the utility set is the baseline-eligible UHR denominator.}
\label{tab:e3-cases}
\end{table}

All 300 D1 records are over-trusted, so COTR is 1.000. Of 300 selected D2 records, 297 are correct under neutral history; severe history induces an error in all 297, so the prespecified denominator gives CIR $297/297=1.000$. The three neutral-history errors are excluded before $\mathcal F_2$ is frozen. The 300-record utility cohort contains 150 both-regular and 150 both-irregular HIGH-reliability cases. The prespecified baseline-eligibility rule freezes $\mathcal H$ after the no-evidence response and before either arm's evidence is scored. All 150 both-regular cases return USE\_AS\_IS and form $\mathcal H$; all 150 both-irregular cases return VERIFY\_FIRST and cannot exhibit the defined transition. E3 UHR is therefore a narrow both-regular endpoint.

\subsection{Evidence donors and inference}

The clear-case evidence pool contains 3,733 records from 180 test patients. Shuffled evidence always uses a different patient and record. Within each Student seed, donor identity is exactly paired between B2 and K2, while each arm renders the corresponding arm- and seed-specific score and label. Donor maps are independently frozen across Student seeds. Mitigation comprises 3,582 shuffled assignments involving 1,476 distinct donor records; utility comprises 1,800 assignments involving 843 distinct donor records. Across both stages, 1,985 distinct donors are used.

Each of 5,000 replicates resamples patients with replacement and then draws three Student seeds with replacement. The same seed draws are shared across arms, matched/shuffled evidence, scenarios, and utility. This preserves paired contrasts while propagating patient and experimental-replicate variation; because donor maps are seed-specific, the seed replicate jointly indexes checkpoint and frozen donor allocation.

\begin{table}[!htbp]
\centering
\setlength{\tabcolsep}{5pt}
\begin{tabular}{llrrr}
\toprule
Scenario & Arm & Matched CRR / UHR & Shuffled CRR & ESRM \\
\midrule
D1 & B2 & .4011 & .1844 & .2167 \\
D1 & K2 & .4483 & .1866 & .2617 \\
D2 & B2 & .9439 & .7856 & .1582 \\
D2 & K2 & .9621 & .7370 & .2251 \\
Utility & B2 & .0289 & --- & --- \\
Utility & K2 & .0356 & --- & --- \\
\bottomrule
\end{tabular}
\caption{E3 pooled absolute rates, averaging the three prespecified Student seeds. Utility entries are UHR on the 150-record baseline-eligible both-regular set. D1 matched K2 CRR of .4483 leaves .5517 of frozen D1 errors unrepaired.}
\label{tab:e3-absolute}
\end{table}

\begin{table}[!htbp]
\centering
\setlength{\tabcolsep}{4pt}
\begin{tabular}{lrrrl}
\toprule
Prespecified K2--B2 endpoint & Estimate & 95\% CI & Criterion & Status \\
\midrule
D1 matched CRR & $+.0472$ & [$+.0167,+.0728$] & Lower $>0$ & Pass \\
D1 ESRM & $+.0450$ & [$+.0098,+.0818$] & Lower $>0$ & Pass \\
D2 matched CRR & $+.0182$ & [$+.0151,+.0291$] & Lower $>0$ & Pass \\
D2 ESRM & $+.0669$ & [$+.0260,+.0719$] & Lower $>0$ & Pass \\
UHR & $+.0067$ & [$-.0040,+.0170$] & Upper $<.020$ & Pass \\
\bottomrule
\end{tabular}
\caption{E3 paired patient-cluster and Student-seed bootstrap. All four efficacy CI lower bounds exceed zero. The UHR CI upper bound is below the exploratory 0.02 interface-engineering limit, which is not clinically validated.}
\label{tab:e3-primary}
\end{table}

\begin{table}[!htbp]
\centering
\setlength{\tabcolsep}{5pt}
\begin{tabular}{lrrrr}
\toprule
K2--B2 effect (percentage points) & Seed 42 & Seed 123 & Seed 2026 & Three-seed mean \\
\midrule
D1 matched CRR & +5.33 & +4.00 & +4.83 & +4.72 \\
D1 ESRM & +4.67 & +6.00 & +2.83 & +4.50 \\
D2 matched CRR & +0.34 & +2.02 & +3.10 & +1.82 \\
D2 ESRM & +6.06 & +6.67 & +7.34 & +6.69 \\
UHR & +0.67 & +0.67 & +0.67 & +0.67 \\
\bottomrule
\end{tabular}
\caption{E3 seedwise effects. Every efficacy effect remains positive across all three Student seeds.}
\label{tab:e3-seedwise}
\end{table}

E3 closes the prespecified internal holdout question. The challenge manipulations again produce frozen errors, and K2 reproduces the validation-stage partial repair and matched-versus-shuffled package-specificity advantages on patient-disjoint test records: all four efficacy intervals exclude zero. D1 matched CRR rises from .4011 to .4483, so most D1 errors remain unrepaired; D2 matched CRR is ceilinged, while ESRM rises from .1582 to .2251. The UHR upper bound meets the exploratory interface limit only on the baseline-eligible both-regular subset. These results support internal replication of conditional behavioral effects under the frozen MIMIC protocol, not natural-distribution net benefit, external transport, comprehensive safety, or clinical outcome improvement.

\section{Exploratory Clinician Evaluation of a Continuous Heart-Rate DFOT Extension}
\label{sec:hr-clinician}

\paragraph{Motivation and task construction.}
The primary study evaluates DFOT using categorical PPG-derived rhythm measurements through the D1/D2 challenges defined in the main paper. As a separate exploratory analysis, we investigated whether the same framework could be instantiated for a continuous derived measurement, namely PPG-derived heart rate.
The categorical-rhythm benchmark and continuous heart-rate extension used separate, record-disjoint subsets of the source cohort.
The HR extension is not part of the primary D1/D2 protocol and does not contribute to its prespecified efficacy or utility gates.

Within the HR extension, the 25 underlying cases were partitioned into two mutually exclusive subsets. The first comprised 12 hidden-tachycardia challenge cases, in which the displayed PPG-derived heart rate was non-tachycardic while the review-only ECG reference indicated tachycardia. An appropriate reliability-aware report should avoid confidently accepting the displayed PPG estimate and recommend verification. The second comprised 13 reliable-normal control cases, in which the PPG-derived and ECG-referenced heart rates agreed and unnecessary verification should therefore be avoided.

Each case generated two reports: one without explicit reliability evidence and one with the language-facing HR reliability interface, yielding 50 reports in total. The no-evidence report was generated without an explicit reliability interface, whereas the evidence-guided report received the categorical HR-reliability label and its associated PPG-only reliability scores. The ECG reference was available only to the evaluator and was never provided to the deployment-time report generator.

\paragraph{Independent clinician scoring.}
An independent clinician evaluated the 50 anonymized reports using a prespecified five-point ordinal scale that jointly considered clinical correctness, safety, and actionability. A score of 5 indicated a fully appropriate report, whereas a score of 1 indicated a clearly incorrect or potentially unsafe report. The clinician was instructed to assess whether the
report appropriately interpreted the displayed PPG-derived heart rate in light of the review-only ECG reference and the accompanying reliability information without rewarding report length or stylistic preferences. The clinician was blinded to GPT-5.5 ratings and internal analysis labels. GPT-5.5 independently scored the same reports using the same rubric.

Analyses were paired by the 25 underlying cases. Mean paired score changes and their 95\% confidence intervals were estimated using 20,000 case-level bootstrap replicates. Clinician--GPT-5.5 concordance was summarized using exact agreement, agreement within one ordinal point, and quadratic-weighted
$\kappa$.
The evaluation involved a single prespecified clinician and was intended as an exploratory face-validity assessment rather than a formal multi-rater validation.

\paragraph{Results.}
For the 12 hidden-tachycardia cases, the clinician's mean score increased from 2.33 without reliability evidence to 4.92 with evidence-guided reporting, corresponding to a mean paired increase of 2.58 points (95\% CI:
$[2.33,\,2.83]$). All 12 cases received higher scores under the evidence-guided condition, indicating that the reliability interface helped the reports avoid confidently accepting an apparently non-tachycardic PPG estimate when the
offline ECG reference indicated tachycardia.

For the 13 reliable-normal controls, the mean score increased from 3.62 to 4.77, with a mean paired increase of 1.15 points (95\% CI: $[0.62,\,1.77]$).Eight cases improved and five were unchanged; none received a lower score. These findings suggest that the higher ratings were not achieved
by indiscriminately recommending ECG verification when the displayed measurement was already reliable.

Clinician and GPT-5.5 ratings agreed exactly for 35 of 50 reports (70\%), and all 50 ratings differed by no more than one ordinal point. Quadratic-weighted agreement
was $\kappa=0.906$. Together, these findings provide exploratory
face-validity evidence that the DFOT evaluation framework can be instantiated for a continuous derived measurement in addition to categorical rhythm.
However, because the evaluation involved a single clinician, a small number of cases, and a composite ordinal score, it should not be interpreted as multi-clinician consensus, probability calibration, prospective clinical
validation, or evidence of improved patient outcomes.

\begin{figure}[t]
    \centering
    \includegraphics[width=\linewidth]
    {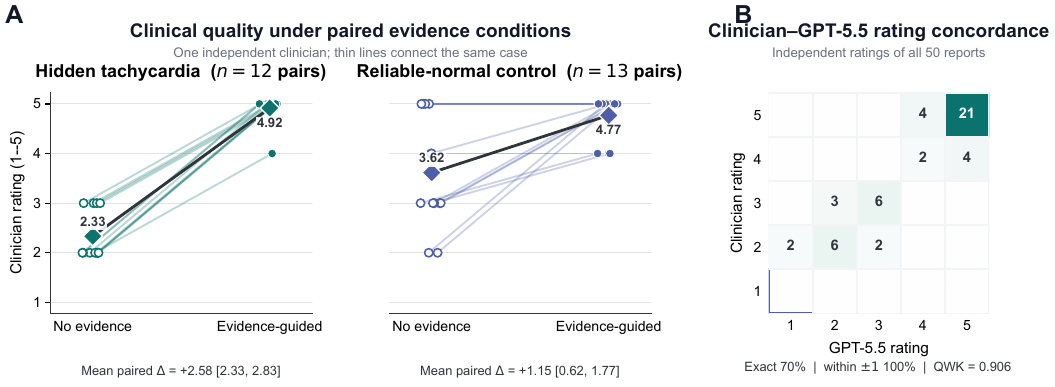}
    \caption{\textbf{Exploratory clinician evaluation of the continuous heart-rate DFOT extension.}
    \textbf{(A)} One clinician, blinded to GPT-5.5 ratings and internal analysis
    labels, scored 50 reports from 25 paired cases on a prespecified 1--5 scale
    integrating clinical correctness, safety, and actionability. Thin lines
    connect reports from the same underlying case, and diamonds indicate group
    means. Bracketed values show 95\% case-bootstrap confidence intervals for
    the mean paired change based on 20,000 replicates.
    \textbf{(B)} Independent clinician--GPT-5.5 rating concordance across all
    reports. Cells show report counts. Exact agreement was 70\%, all ratings
    differed by at most one point, and quadratic-weighted agreement was
    $\kappa=0.906$.}
    \label{fig:hrd1-clinician-evaluation}
\end{figure}

\section{Tail Mechanism Audit}

Global ranking metrics do not indicate where improvements occur along the ROC curve and may therefore overlook behavior in the stringent low-false-positive operating region. We therefore explored a tail-aware objective on the development set under a prespecified stop rule. To determine whether any apparent gain depended on label-specific privileged information rather than generic optimization effects, we designed a negative control that preserved the number of positive examples and all training mechanics while disrupting teacher--student label correspondence. The final patient-deranged shuffled-tail control uses patient-disjoint donors, preserves all 346 positive assignments, and retains the true hidden label for only 10.69\% of reassigned examples.

\begin{table}[!htbp]
\centering
\setlength{\tabcolsep}{4pt}
\begin{tabular}{lrrrr}
\toprule
Comparison & $\Delta$ tail mean & 95\% CI & $\Delta$ standardized pAUC & 95\% CI \\
\midrule
True-tail -- K2 anchor & +3.06 pp & [1.28,5.56] & +.0138 & [.0067,.0240] \\
True-tail -- hidden-weighted BCE & +2.59 pp & [.85,4.53] & +.0120 & [.0038,.0212] \\
True-tail -- patient-deranged shuffle & +1.31 pp & [-1.70,4.50] & +.0057 & [-.0105,.0212] \\
K2 true-tail -- B2 true-tail & +2.30 pp & [-1.06,6.51] & +.0108 & [-.0059,.0295] \\
\bottomrule
\end{tabular}
\caption{Development-only tail confirmation. The effective shuffled-tail comparison crosses zero, so label specificity is not demonstrated.}
\label{tab:tail}
\end{table}

The frozen decision is \texttt{STOP\_TAIL\_GAIN\_NOT\_LABEL\_SPECIFIC}: no full-data tail escalation, no test-tail access, and no additional tuning of margin, weight, or hard-negative ratio. This means ``label specificity was not demonstrated,'' not ``the objective has no effect.''

\section{Protocol Design Lessons and Corrective Actions}

\subsection{No-evidence leakage}

An early baseline prompt explicitly instructed the LLM to verify when reliability evidence was absent. The model consequently avoided DFOT before characterization, producing no induced subset. Although the run completed technically, CRR and ESRM were undefined and the responses were excluded.

\subsection{Clinical-status/reliability collision}

A second protocol used one TRUSTWORTHY/SUSPECT output for both measurement reliability and rhythm status. The model interpreted SUSPECT clinically, while the scorer interpreted it as unreliability. The gate observed no valid D1/D2 characterization and stopped the evidence stage before tens of thousands of invalid calls.

\subsection{Resulting rules}

These failures produced six requirements: characterize before mitigating; use mechanism-specific D1/D2 prompts; separate measurement use from clinical status; randomize only the donor source while holding the case and evidence template fixed (the resulting label and score may change jointly); reuse the exact baseline answer in reactive mitigation; and freeze prompt, parser, cases, thresholds, and statistics together.

\section{Claim Registry}
Table \ref{tab:claims} summarizes the E0--E3 frozen claim registry and the corresponding scope of interpretation.


\begin{table*}[t]
\centering
\small
\setlength{\tabcolsep}{4pt}
\renewcommand{\arraystretch}{1.15}
\begin{tabular}{p{1.45in}p{0.95in}p{2.15in}p{2.10in}}
\toprule
\textbf{Claim} &
\textbf{Status} &
\textbf{Supported conclusion} &
\textbf{Do not conclude} \\
\midrule

\multicolumn{4}{l}{\textbf{Framework}}\\
\midrule
DFOT formulation &
Introduced &
Names a downstream failure caused by inappropriate reliance on derived measurements. &
Equivalent to hallucination or upstream misclassification. \\

Metric chain &
Introduced &
Five estimands separate induction, repair, specificity, and interface harm. &
A single metric fully characterizes DFOT. \\

Matched--shuffled intervention &
Validated &
Tests whether downstream revision depends on instance-specific evidence. &
Identifies the separate effects of label, score, or wording. \\

\midrule
\multicolumn{4}{l}{\textbf{Upstream evidence generation}}\\
\midrule

Upstream transfer &
Confirmed &
K2 improves PPG-only reliability ranking over B2. &
PPG reconstructs ECG or ECG is ground truth. \\

Matched-target specificity &
Confirmed &
Supports an instance-correspondence component beyond generic regularization. &
The entire gain is explained by recovered ECG semantics. \\

Extreme low-FPR transfer &
Not confirmed &
Global ranking and low-FPR behavior remain distinct. &
All safety operating points improve. \\

\midrule
\multicolumn{4}{l}{\textbf{Downstream DFOT evaluation}}\\
\midrule

D1 repair &
Confirmed &
K2 improves matched repair for current-measurement conflicts. &
Clinical-error reduction or external transport. \\

D1 specificity &
Confirmed &
K2 improves D1 ESRM for the matched evidence package. &
ESRM isolates label, score, or wording effects. \\

D2 repair &
Confirmed &
K2 improves matched repair under misleading history. &
Real-world longitudinal robustness. \\

D2 specificity &
Confirmed &
K2 improves D2 ESRM despite near-ceiling CRR. &
ESRM explains all context conflicts or internal LLM mechanisms. \\

Interface harm (UHR) &
Criterion met &
The exploratory engineering limit is satisfied on the baseline-eligible utility set. &
Clinical non-inferiority, absence of harm, or validated utility. \\

\midrule
\multicolumn{4}{l}{\textbf{Robustness}}\\
\midrule

Cross-LLM robustness &
Supported &
Positive effects persist across the three evaluated LLMs. &
Universal robustness across LLM families. \\

Prompt robustness &
Supported &
Positive effects persist across P1--P3. &
Invariance to arbitrary prompts or model--prompt interactions. \\

E3 replication &
Confirmed &
All four efficacy gates replicate on held-out MIMIC patients. &
External-dataset or prospective transport. \\

\midrule
\multicolumn{4}{l}{\textbf{Clinical interpretation}}\\
\midrule

Clinical outcome &
Not tested &
Results concern controlled downstream behavior. &
Clinical safety or improved patient outcomes. \\

\bottomrule
\end{tabular}
\caption{Frozen claim registry summarizing the scope of the supported conclusions.}
\label{tab:claims}
\end{table*}

\section{Reproducibility Checklist for Authors}

E0--E3 freeze the locked Student results, validation D1/D2 effects, cross-LLM and cross-prompt robustness endpoints, the patient-disjoint downstream test replication, the exploratory UHR interface criterion, and the claim registry. The E3 protocol records the case and donor manifests, synchronized bootstrap design, exact Qwen settings, deterministic parser, and input/output hashes. The executable release should include preprocessing and patient-split builders, case/donor manifests or an authorized-user reconstruction route under the MIMIC data-use agreement, full prompt/parser artifacts, and checkpoint/configuration hashes. The following items remain unavailable or require author verification:
\begin{itemize}
    \item institutional ethics approval, consent/waiver, and data-use permissions;
    \item device/vendor metadata and formal-cohort age, sex, and race statistics, which are absent from the frozen waveform artifacts;
    \item the three teacher-output class names/index order, their label-generation rule, teacher/student target correspondence, and whether window/global classifier parameters are shared;
    \item a full Qwen checkpoint revision or complete shard hashes, the exact DeepSeek/GPT service revisions, and non-vLLM runtime packages;
    \item ECG-teacher wall-clock time and complete token/cost ledger;
    \item final data/code release level and a MIMIC-compliant reproducible route for authorized reviewers.
\end{itemize}

\end{document}